\def\year{2018}\relax
\documentclass[letterpaper]{article} 
\usepackage{aaai18}  
\usepackage{times}  
\usepackage{helvet}  
\usepackage{courier}  
\usepackage{url}  
\usepackage{graphicx}  
\usepackage{amsmath}
\usepackage{algorithm}
\usepackage[noend]{algpseudocode}
\begin{document}

\title{An Automated Question-Answering Framework Based on \\
Evolution Algorithm}

\author{Sinan Tan\textsuperscript{1},  Hui Xue\textsuperscript{2},  Qiyu Ren\textsuperscript{3},  Huaping Liu\textsuperscript{1} \and  Jing Bai\textsuperscript{4} \\
\textsuperscript{1}{Tsinghua University, China}\\
\textsuperscript{2}{Microsoft Research, China}\\
\textsuperscript{3}{University of Posts and Telecommunications, China}\\
\textsuperscript{4}{Microsoft, Silicon Valley, U.S}\\
}

\maketitle

\begin{abstract}
Building a deep learning model for a Question-Answering (QA) task requires a lot of human effort, it may need several months to carefully tune various model architectures and find a best one. It’s even harder to find different excellent models for multiple datasets. Recent works show that the best model structure is related to the dataset used, and one single model cannot adapt to all tasks.
In this paper, we propose an automated Question-Answering framework, which could automatically adjust network architecture for multiple datasets. 
Our framework is based on an innovative evolution algorithm, which is stable and suitable for multiple dataset scenario. 
The evolution algorithm for search combine prior knowledge into initial population and use a performance estimator to avoid inefficient mutation by predicting the performance of candidate model architecture. 
The prior knowledge used in initial population could improve the final result of the evolution algorithm. The performance estimator could quickly filter out models with bad performance in population as the number of trials increases, to speed up the convergence.
Our framework achieves 78.9 EM and 86.1 F1 on SQuAD 1.1, 69.9 EM and 72.5 F1 on SQuAD 2.0. On NewsQA dataset, the found model achieves 47.0 EM and 62.9 F1.
\end{abstract}

\section{1. Introduction}
Question-Answering (QA) is a key problem in the field of artificial intelligence.
It requires one to find the correct answer for a given question from passages.
In our work, we focus on the task of Reading Comprehension,
where the question is grounded on related documents or passages.
Due to the diversity of language expression and the complexity of grammar,
understanding the semantic of long passages and queries has always been a difficult task.
Recently, a large number of deep learning models including BiDAF \cite{seo2016bidirectional}, R-Net \cite{wang2017gated}, FusionNet \cite{huang2017fusionnet} and QANet\cite{yu2018qanet} are proposed by researchers,
and they do achieved good performance on certain datasets (e.g. SQuAD dataset\cite{rajpurkar2016squad}).

However, one single model cannot adapt to all the tasks.
For example, \cite{joshi2017triviaQA} shows that a model that performs well for SQuAD may fail to get outstanding result on TriviaQA.
Therefore, for different reading comprehension datasets, different models may need to be designed, which will take a lot of human effort and time.

In this work, we propose a new evolution algorithm based method to build an automated Question-Answering framework.
Leveraging neural architecture search (NAS) and transfer learning,
we are able to design suitable model architectures for different reading comprehension tasks, requiring significantly less time and human effort.

Most existing algorithms \cite{zoph2016neural,Zoph2017TNAS,pham2018efficient,Liu2018DART,tan2018mnasnet} for neural architecture search, are optimized for a predefined model with strong knowledge of domain experts. They focus on designing a directed acyclic graph (DAG) as a cell structure for the model, and transfer their knowledge between different tasks by sharing and repeating same basic cell over and over again. However, from the previous work\cite{seo2016bidirectional,wang2017gated,huang2017fusionnet,yu2018qanet}, most state-of-art models in reading comprehension consists of a variety of complex connections, and do not contain any largely repeated model structure. Therefore it is almost impossible to design a predefined global structure for reading comprehension models.

In addition, the existing methods like ENAS\cite{pham2018efficient}, based on reinforcement learning and using s deep learning model as a controller, are not suitable for finding model architectures for multiple datasets. Due to the instability of reinforcement learning\cite{mnih2015human}, different datasets may require different hyper-parameters, which means human intervention is actually necessary. So it's hard to apply these methods for various datasets.

We build a flexible framework to search for Reading Comprehension models. We do not predefine any high-level structure for the models. In other words, model architectures can grow dynamically in our framework. Besides, our method, which uses evolution algorithm, is insensitive to hyper-parameters. It is very suitable for the scenario of multiple datasets.

Previous research work\cite{Real2017Evolution,Jaderberg2017Population,real2018regularized} have shown that evolutionary algorithm are feasible to find a neural network architecture for image classification. Our method is also based on evolution algorithm. Compared to image recognition, the Reading Comprehension task is very different. The models in Reading Comprehension usually need to use Recurrent Neural Networks(RNN) layers to encode the semantics and attention layers to compute the relationship between query and document, so the models  suitable for Reading Comprehension are more complex and have longer training time. This means the search time of evolution might be very long. 

In order to solve this problem, we propose two main ideas.
First, the models used to initialize population in our evolution algorithm come from the state-of-art models and diverse stochastic models mutated from those models.
In traditional evolution algorithms\cite{Real2017Evolution}, a large number of experiments in the early stages are needed to explore and get a better direction of mutation, which is very inefficient.
Searching from the state-of-art models could introduce human knowledge into the search progress and improve the final result in limited time, which could be concluded from our experiments.
Second, we use a performance estimator to estimate the performance of candidate models and try to model the relationship between performance and model architectures. The concept of using neural network as performance predictor has been proposed in Peephole \cite{deng2017peephole}.
When performing the next mutation, the performance estimator will be used to evaluate the performance of several different possible mutated models and our framework could pick best candidate to train. This way, we can avoid inefficient random mutation, and speed up the search progress. 

Here is the summarization of our contributions in the proposed framework:

\begin{itemize}
\item We build an automated Question-Answering framework based on innovative evolution algorithm. It’s a flexible framework and can automatically adapt to different Reading Comprehension datasets and produce competitive results. 
\item We propose an improvement based on naive evolution algorithm\cite{Real2017Evolution}, which is initializing population from the state-of-art models. This can improve the performance of the found model.
\item We use a CNN-based performance predictor to estimate the performance of candidate mutations, which is a more efficient alternative to naive random mutation. This can accelerate the search progress and speed up its convergence.
\end{itemize}

\section{2. Related Work}
Our work is highly related to neural architecture search (NAS) and Reading Comprehension. Here we will summarize previous work in the two fields.

\subsection{2.1 Neural Architecture Search.} 
NAS\cite{zoph2016neural} uses a RNN controller to discover neural network architectures by searching for an optimal graph to maximize the expected accuracy of the generated architectures on a validation set.
NAS takes a lot of time for getting a single model.
Besides, ENAS\cite{pham2018efficient} speeds up the search progress by sharing parameters among the models being searched. \cite{cai2018efficient} use Net2Net transfer \cite{chen2015net2net} to speed up the search when updating the network structure.

But these methods are not suitable for the multiple datasets scenario in reading comprehension for two reasons. 
First, the methods based on reinforcement learning or deep learning models, are very sensitive with their hyper-parameters and thus aren't) suitable for the multiple dataset scenario.
Second, unlike image recognition, known Question-Answering models are very complex. Most competitive QA models do not contains a large number of repeated basic blocks similar to ResNet \cite{he2016deep}. Therefore, it's impossible to find a cell structure suitable for every dataset. Our framework is inspired by\cite{Real2017Evolution}, which proposes a paralleled evolution algorithm, by randomly picks two models from the population and randomly mutates the better one. But there are two drawbacks in original evolution. One is random mutation, which is ineffective and uncontrollable. The other is that the evolution algorithm consumes a large amount of resources to filter the weak individual compared to other methods. Our method uses known state-of-art models to initialize the population, and uses performance estimator to avoid inefficient mutation and speed up the convergence.

\subsection{2.2 Reading Comprehension.} 

\begin{figure*}[h]
    \centering
    \includegraphics[scale=0.35]{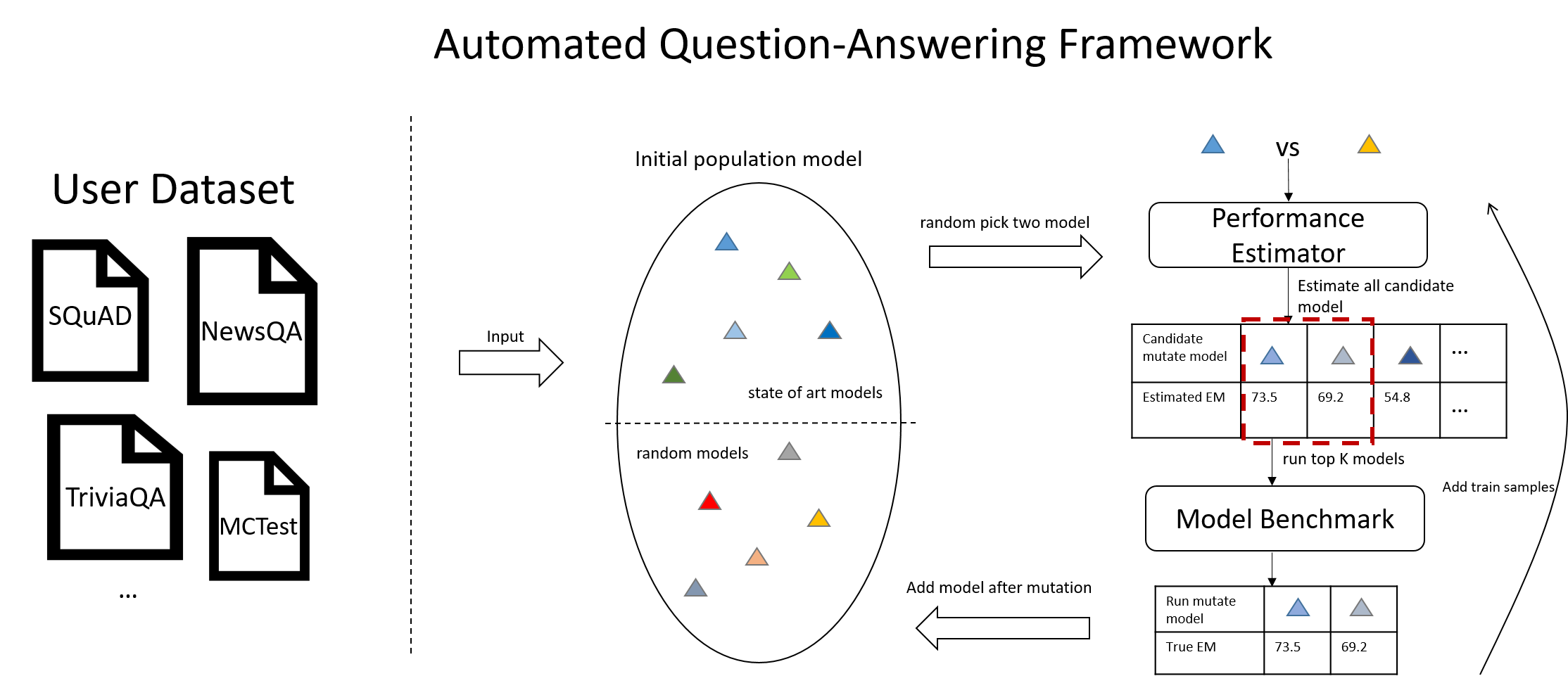}
    \caption{Overall architecture of our framework}
\end{figure*}

There are many systems for reading comprehension that employ embedding at the character level and word level. \cite{chen2017reading} applied some linguistic features to input embedding. \cite{peters2018deep} introduce a new word embedding called ELMo, which is learned by training a bidirectional language model. \cite{mccann2017learned} use a deep LSTM encoder trained for machine translation as contextualized word vectors called CoVe. Adding ELMo and CoVe could improve the performance of the common NLP tasks, by using information from large datasets. In our paper, we take advantage of these embeddings as the input layer in our automated QA framework. \cite{wang2016machine} propose match-LSTM to predict answer boundaries with pointer network. The pointer network has also become a common method for the model of reading comprehension. We use pointer-network as the output layer for predicting the answer span. \cite{seo2016bidirectional} use use bi-directional attention flow mechanism to obtain a query-aware context representation. \cite{wang2017gated} proposed a self-matching attention mechanism to refine the representation by matching the passage against itself, which effectively encodes information from the whole passage. \cite{huang2017fusionnet} proposed a fully-aware multi-level attention mechanism to capture the complete information in one text (such as a question) and exploit it in its counterpart (such as context or passage) layer by layer. \cite{yu2018qanet} proposed use convolution and self attention to encode text, where convolution models local interactions and self-attention models global interactions.

These papers focus on modeling the interactions between the context and the query. But for different datasets, the interaction may complex and have no fixed pattern. \cite{joshi2017triviaQA} showed that BiDAF performs well in SQuAD 1.1 but failed in TriviaQA, which means different tasks need different model. In our paper, we leverage these research works and use those proposed layers to design the search space for QA model. As for the connections between those layers, we use the proposed evolution algorithm based neural architecture search to find the best solution.

\section{3. An Automated Question-Answering Framework}
Figure 1 shows the pipeline of our Question-Answering framework. Here is an overview of our proposed framework:
Before being fed into the model, all documents and questions in a dataset will be tokenized and prepossessed to extract linguistic features.
Then the state-of-art of model architecture in QA and diverse random model architectures mutated from them will be added into initial populations.
Following the traditional evolution algorithm\cite{Real2017Evolution}, the framework will randomly pick two individual(models) from populations.
We then choose the model that achieves better performance before mutation.
We propose an optional variant of the traditional mutation as an alternative to random mutation. It uses a CNN-based performance predictor to estimate the performance of models after possible mutations.
We will keep the history of the model structures having been trained in the process of evolution,
and train the performance prediction model with the structure of those QA models and their performance.

In the following sections, we will explain details for different components in our framework.

\subsection{3.1 Initial Population} 
For better use of prior knowledge, we consider adding state-of-art models into initialized population,
including BiDAF \cite{seo2016bidirectional}, RNET \cite{wang2017gated} and FusionNet \cite{huang2017fusionnet} and so on.
It is also possible to add even more known models to the framework, but we are going to stick with those three models in our experiments.

Models designed by humans beings could be considered as a local optima for neural architecture structure.
Using them speeds search but may also limit our search space.
Therefore, other models, obtained by performing a large number of mutations on those known models are also added to the initial population, which could help to guarantee the diversity of population. 

\subsection{3.2 Search Space}
Table 1 shows the statistics about the structures of other state-of-art models.
Based on that result, we use the following mutation actions in our proposed framework.

\begin{itemize}
\item IDENTITY (Effectively means “keep training”).
\item INSERT-RNN-LAYER (Inserts a LSTM. Comparing the performance of GRU and LSTM in our experiment, we decided to use LSTM here.)
\item REMOVE-RNN-LAYER
\item INSERT-ATTENTION-LAYER(Inserts a attention layer. Comparing with other attention, we consider use symmetric attention proposed by FusionNet \cite{huang2017fusionnet})
\item REMOVE-ATTENTION-LAYER
\item ADD-SKIP (Identity between random layers).
\item REMOVE-SKIP (Removes random skip).
\item CONCAT-TWO-INPUT(Combines output of two layers into one using concat operation)
\end{itemize}

The layers contained in search space were chosen for their similarity to the actions that a human designer may take when improving an architecture.
The probabilities for these mutations are equal before they are estimated by the performance estimator.

\begin{figure}
    \centering
    \includegraphics[scale=0.3]{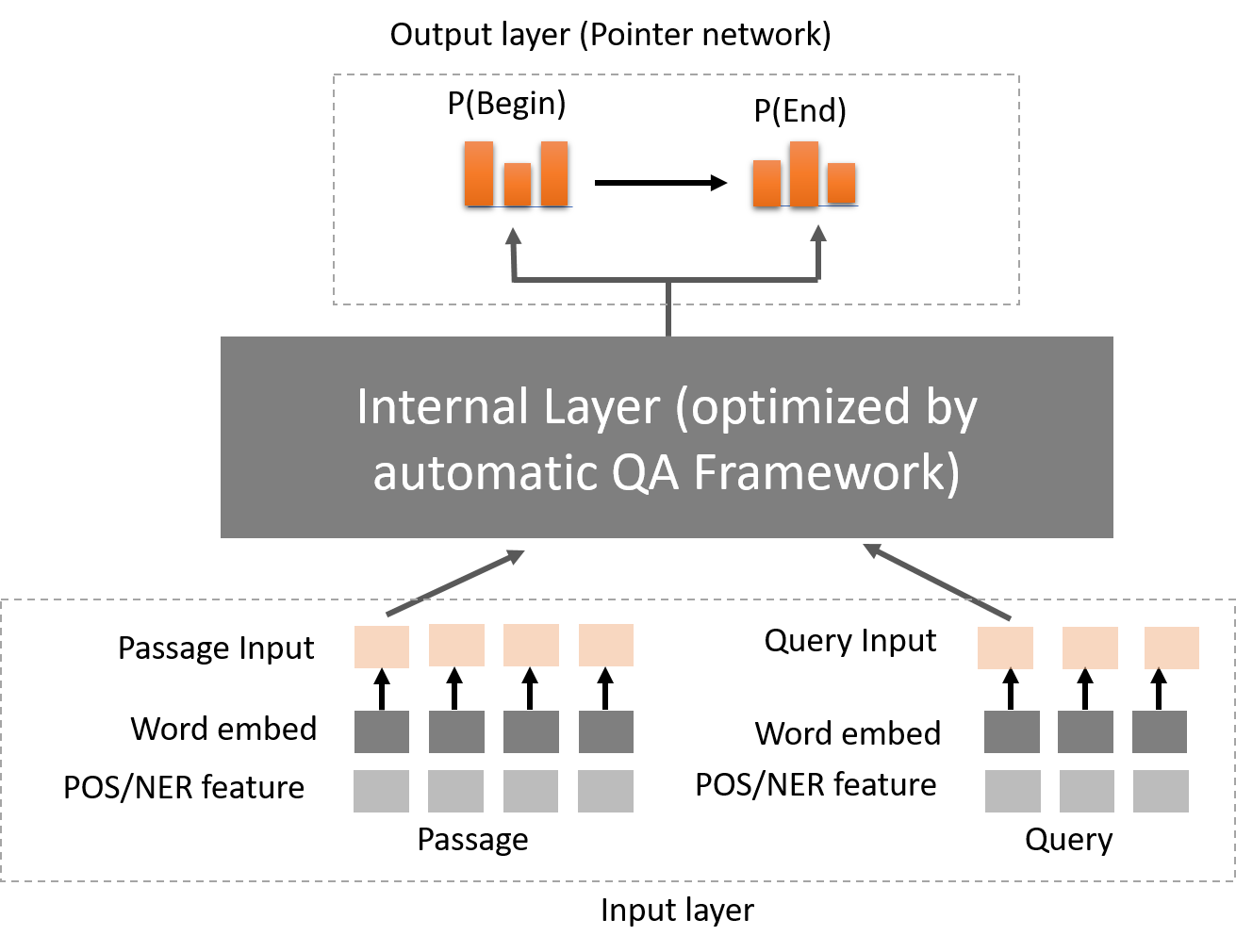}
    \caption{Skeleton of our models}
\end{figure}

\subsection{3.3 Performance Estimator}

\begin{table*}
    \begin{center}
    \begin{tabular}{|l|c|c|c|c|c|}
    \hline \bf Model Name & \bf RNN & \bf Attention & \bf Skip Connection & \bf Conv & \bf Concat \\ \hline
    BiDAF\cite{seo2016bidirectional} & Yes & Yes & No & No & No \\
    DCN\cite{xiong2016dynamic} & Yes & Yes & No & No & Yes \\
    ReasoNet\cite{shen2017reasonet} & Yes & Yes & No & No & No \\
    R-net\cite{wang2017gated} & Yes & Yes & No & No & Yes \\
    FusionNet\cite{huang2017fusionnet} & No & Yes & Yes & Yes & Yes \\
    QANet\cite{yu2018qanet} & Yes & Yes & No & No & No \\
    \hline
    \end{tabular}
    \end{center}
    \caption{\label{tab} Building blocks of known QA models.}
\end{table*}

In the search progress, we will generate several valid candidate models with the mutation space mentioned above. 
And then our framework will use the performance estimator to estimate the EM score of each candidate model.

To estimate the score of each candidate model, here we use CNN instead of LSTM to build the prediction model. The model configuration will be represented by an adjacency matrix of the computation graph. We believe that CNN is able to capture shared structures in the neural network, as those structures will exhibit locality in adjacency matrix.

An adjacency matrix can be embedded to an 2D feature map, which have been proven can be effectively processed by convolution neural network. Each adjacency matrix's shape is $N \times N \times K$, where $N$ is the maximum number of nodes the model may contain,
and a K-dimensional embedding is used to represent the relation of nodes.

The embedding of relation for the layers are related to the search space, for example, if node $X$ is the output of another node $Y$,
after getting through an LSTM layer, then the cell in the $X$-th row and $Y$-th column of the feature map will be the embedding vector of the relation
``is input of LSTM``,
and the cell in the $Y$-th row and $X$-th column of the feature map will be marked to the embedding vector of the relation
``is output of LSTM``. The way of encoding attention layer, concatenate layer, and the skip connections is also similar to this.
``No connection``, ``Self``, ``Padding`` are considered as special relation types, and are also added to the feature map matrix.

We use a PreResNet50 \cite{he2016identity} as the model to predict the performance of the reading comprehension model in our framework.
The output layer is changed to be normalized by sigmoid function: $\frac {1} {1 + e^{-x}}$, representing the exact match score.

We use the following L2 loss for the training procedure.

\begin{equation}
    L(\theta) = (r - r^*)^2
\end{equation}

where $\theta$ is parameters of the performance estimator, $r$ is the predicted performance of the model, $r*$ is the actual performance of the model.

During the search progress, an increasing number of models will be trained, which means more and more training data will be available for the performance estimator.
Therefore, the probability ($p$) of the model with better predicted performance chosen will change by time:

\begin{equation}
  p = 0.5 + 0.5 * \frac {min(epoch, epoch_{max})} {epoch_{max}}
\end{equation}

Where $epoch_{max}$ is a constant after which number of epochs model with better predicted performance will always be chosen, $epoch$ is the number of model trained. The intuition is that, as the number of models being trained increasing, the performance estimator become more and more reliable. So as $epoch$ is increasing, $epoch$ is also expected to increase, the probability of adopting the result proposed by the performance estimator is higher.

The use of this performance estimator is optional. We analyze how it affects the training procedure in the last section of the paper.

\subsection{3.4 Model Configuration}

The model configuration in our framework contains three parts: input layer, internal architecture, and output layer. The input layer and output layer are fixed in our framework. internal architecture is only part we try to tune. Model Benchmark is a toolkit that will parse these configurations and compile them to a model that could run in a deep learning platform, after which we get the performance of this model.

\begin{itemize}
\item Input Layer

For the word embedding, we concatenate the pre-trained 300-dimensional GloVe vectors \cite{pennington2014glove},
1024-dimensional ELMo vectors \cite{peters2018deep} and 600-dimensional CoVe vectors \cite{mccann2017learned}. Following \cite{chen2017reading}, we use three additional types of linguistic features for each word : 1) a 44-dimensional POS tagging embedding, 2) an 13-dimensional NER tagging embedding, and 3) a 3-dimensional binary exact match feature. The character level embedding is the same as\cite{wang2017gated}, taking the final hidden states of a bi-directional RNN applied to embedding of characters in the token. We concatenate all these embeddings as input layer in our framework for queries and documents.

\item Internal Architectures

Internal Architectures is the key part in our framework.
The internal architectures is generated from the configuration and in the process of evolution.

The layers in internal architecture are always from the search space.
    
\begin{itemize}
\item LSTM

LSTM\cite{hochreiter1997long} is a common type of RNN widely used in language models. In our experiments, models with LSTM outperform those with GRU \cite{cho2014learning}. So we use LSTM instead of GRU to extract language features.

\item Attention

We use the symmetric attention function proposed by \cite{huang2017fusionnet}, where the scoring function is

\begin{equation}
    s = f(U x)^T D f(U y)^T
\end{equation}

We replaced the non-linearity function $ f $ with Swish activation function \cite{ramachandran2017swish}, because we found that Swish activation speeds the training process of a single model.

The above attention scoring function is used for both self-attention and document-to-question or question-to-document attention.

\item Concatenate Layer

Following \cite{huang2017fusionnet}, we use the concatenate layer to fuse different level of concepts is helpful for the model. Here we simply introduce this operation to our framework, which concatenate the feature dimension of document or question vector.

\item Skip-Connection Layer

Skip connection is very common in very deep convolution neural networks, and is used by \cite{yu2018qanet}. Whether or not it works for RNN based architecture is still uncertain, but considering the potential benefit, we still add it to our framework.

\end{itemize}

\item Output layer

We follow \cite{wang2017gated} and use pointer networks \cite{wang2016machine} to predict the start and end positions of the answer. We use an avg-pooling over the query representation to generate $u^Q$ for the pointer network.
\begin{equation}
    u^Q = avg\_pool(\{ u^Q_t \}^{n}_{t=1})
\end{equation}
Then we attend for the span start using the summarized question understanding vector $u^Q$.

\begin{align}
\begin{split}
    s^t_i &= V^S tanh(W^S u^P_i + W^Q u^Q) \\
    P^S_i &=  \frac {exp(s^t_i)} {\sum_{i=1}^{n}exp(s^t_j)}
\end{split}
\end{align}

To use the information of the span start, we combine the context understanding vector for the span start with $u^Q$ through a GRU cell \cite{cho2014learning}.

\begin{align}
\begin{split}
    v^Q &= GRU(u^Q, \sum_{i=1}^{n}P^S_i u^P_i)\\
    e^t_i &= V^E tanh(W^E u^P_i + W^Q u^Q) \\
    P^E_i &= \frac {exp(e^t_i)}  {\sum_{i=1}^{n}exp(e^t_j)}
\end{split}
\end{align}

During training, we minimize the sum of the negative log probabilities of the ground truth start and end positions in the predicted distributions. During prediction, we keep the answer span $pos^s,pos^e$ with maximum $P^S_{pos^s}P^E_{pos^e}$ and $pos^s \le pos^e - 15$. For datasets with questions impossible to answer, a special ``no answer`` vector is append to the document sequence. When predicted $pos^s$ points to this vector, the problem is considered unanswerable.
\end{itemize}

The Pointer Network is always used as the output layer in our framework.
Figure 2 shows the skeleton of our model.

\subsection{3.5 Algorithm Detail}

\begin{table*}
    \begin{center}
    \begin{tabular}{|l|c|c|}
    \hline \bf Single Model & \bf Exact Match & \bf F1 \\ \hline
    FastQA\cite{weissenborn2017making} & 68.4 & 77.1 \\
    BiDAF\cite{seo2016bidirectional} & 68.0 & 77.3 \\
    SEDT\cite{liu2017stochastic} & 68.2 & 77.5 \\
    RaSoR\cite{liu2017structural} & 70.8 & 78.7 \\
    FastQAExt\cite{weissenborn2017making} & 70.8 & 78.9 \\
    ReasoNet\cite{shen2017reasonet} & 70.6 & 79.4 \\
    DrQA\cite{chen2017reading} & 70.7 & 79.4 \\
    R-net\cite{wang2017gated} & 75.7 & 83.5 \\
    FusionNet\cite{huang2017fusionnet} & 75.3 & 83.6 \\
    QANet\cite{yu2018qanet} & 76.2 & 84.6 \\
    \bfseries{EvolutionRC} & \bfseries{78.9} & \bfseries{86.1} \\
    R.M-Reader\cite{hu2017reinforced} & 78.9 & 86.3 \\
    SLQA+\cite{wang2018multi} & 80.0 & 87.0 \\
    \hline
    \end{tabular}
    \end{center}
    \caption{\label{tab} The performance of EvolutionRC and competing models on SQuAD 1.1 development set.}
\end{table*}

After the above discussion about the details of our framework, we show the modified version of evolution algorithm used in our framework.

\begin{algorithm}[h]
\caption{Evolution-based model search with performance estimator}\label{euclid}
\begin{algorithmic}[1]
\State $\textit{N} \gets \textit{population size}$
\State $\textit{PI} \gets \textit{existing models}$
\State $\textit{M} \gets \textit{performance estimator update frequency}$
\State $P \gets PI$
\For {$i \in 0:(N-len(PI))$}
    \State $PI\_sample \gets PI.random\_choice()$
    \State $P \gets P + PI\_sample.multiple\_mutate()$
\EndFor
\For {$w \in workers$}
    \State $i, j = sample(P)$
    \If {$performance(i) > performance(j)$}
        \State Delete \textit{j} from P
        \State $k \gets mutate(i)$
    \Else
        \State Delete \textit{i} from P
        \State $k \gets mutate(j)$
    \EndIf
    \State $p = 0.5 + 0.5 * \frac {min(epoch, epoch_{max})} {epoch_{max}}$
    \If {$random(0.0, 1.0) > p$}
        \State $k \gets k.random\_choice()$
    \Else
        \State $k \gets k.choice\_with\_value\_function()$
    \EndIf
    \If {$total\_models\_trained() \% M == 0$}
        \State Update performance estimator model
    \EndIf
    \State Train $k$ and put it into population
\EndFor
\end{algorithmic}
\end{algorithm}

\section{4. Experiments}

We test our framework on three different datasets: NewsQA \cite{trischler2016newsqa}, SQuAD 1.1 \cite{rajpurkar2016squad}, SQuAD 2.0 \cite{rajpurkar2018know}.

Unlike other existing datasets like MSMARCO \cite{nguyen2016ms},these datasets focus exclusively on questions with answers that can be found in the document. NewsQA is a Reading Comprehension dataset containing over 100,000 human-written question-answer pairs, whose documents are from CNN/Daily Mail. Comparing with SQuAD 1.1, the number of question-answer pairs of NewsQA is slightly smaller but the average length of query and documents is much longer. SQuAD 2.0 is much more harder dataset with over 50,000 unanswerable questions written adversarially by crowd-workers. Those unanswerable questions are produced by looking up similar answerable ones. \cite{rajpurkar2018know} shows a strong neural model for SQuAD 1.1 dataset suffers from a huge drop when transfer to SQuAD 2.0.

\subsection{4.1 Experiment Setup}

\begin{table*}
    \begin{center}
    \begin{tabular}{|l|c|c|}
    \hline \bf Single Model & \bf Exact Match & \bf F1 \\ \hline
    BNA\cite{rajpurkar2018know} & 59.8 & 62.6\\
    DocQA\cite{rajpurkar2018know} & 61.9 & 64.8 \\
    DocQA + ELMo\cite{rajpurkar2018know} & 65.1 & 67.6 \\
    \bfseries{EvolutionRC} & \bfseries{69.9} & \bfseries{72.5} \\
    \hline
    \end{tabular}
    \end{center}
    \caption{\label{tab} The performance of EvolutionRC and competing models on SQuAD 2.0 development set.}
\end{table*}

For the evolutionary algorithm, we set the initial population size to 32.
The maximum number of layers is limited to 50.
The number of workers is 8, which means we run the architecture search in 8 GPUs in parallel. In these experiments, we do not use the performance estimator to speed up the training procedure.

For the SQuAD v1.1 and SQuAD v2.0 datasets the GPU we used is Tesla P100-PCIE.
For NewsQA dataset, whose average context length is much longer, we use Tesla P40-PCIE, which has 24GB of RAM in total.

For data preprocessing, we use the tokenizer from the Stanford CoreNLP \cite{manning2014stanford} to preprocess each passage and question and to extract part-of-speech tagging (POS) and named entity recognition (NER) features.
We also apply dropout \cite{srivastava2014dropout} before all RNN layers, after embedding layers, and before linear transform in attention cell.
The dropout rate for RNN and embedding layers is 0.5. The dropout rate for attention cell is 0.3.

The model is optimized by Adamax \cite{kingma2014adam}, with initial learning rate $lr=2 \times 10 ^ {-3}$, $\beta=(0.9, 0.999)$, and $\epsilon=1 \times 10 ^ {-8}$.

We train each model for 50 epochs. Also we use early stopping to avoid overfitting during the trial experiments. The max patience is 5 epoch.
In addition, the patience will increase when the performance is still increasing with numbers very closed to patience of epochs are trained.

For each dataset, we stop the experiment after 500 models are trained. This costs about one week for 8 GPUS, or about 50 GPU days. However, we believe that it is possible to use lower dropout rate and downsampled datasets to significantly lower the computation cost.

The batch size is 64 for the SQuAD v1.1 and SQuAD v2.0 datasets, and is 32 for NewsQA dataset.
When searching for the architecture, we only use CoVe feature.
However, we use both CoVe and ELMo feature when retraining the found model.

We use the official evaluation script for all datasets. We use \cite{NNI} as our experiment platform.

\subsection{4.2 Main Results}

Using our proposed framework for reading comprehension, we performed architecture search on SQuAD 1.1, SQuAD 2.0, and NewsQA Dataset.
We report average exact match and F1 scores for our model after retrained with ELMo embedding.
The searched model achieves 78.9 EM and 86.1 F1 on SQuAD 1.1, 69.9 EM and 72.5 F1 on SQuAD 2.0.
On NewsQA dataset, the found model achieves 47.0 EM and 62.9 F1.
We also compare the model searched by our method with other models designed by humans in Table 2, Table 3, and Table 4.

For SQuAD 1.1 and SQuAD 2.0, our best model is able to outperform many published state-of-art models. But the searched model still cannot reach the performance of top models in the SQuAD competition leaderboard.

We believe that those model might be using more domain knowledge, external data sources, more techniques for data augmentation,
or different ways of predicting the range for the answers, which is beyond the scope for our proposed framework. Besides, we didn't tune hyper-parameters for our best model.

\begin{table*}
    \begin{center}
    \begin{tabular}{|l|c|c|}
    \hline \bf Single Model & \bf Exact Match & \bf F1 \\ \hline
    Match-LSTM\cite{wang2016machine} & 34.4 & 49.6\\
    BARB\cite{trischler2016newsqa} & 36.1 & 49.6 \\
    BiDAF \cite{seo2016bidirectional} & 37.1 & 52.3 \\
    FastQA\cite{weissenborn2017making} & 43.7 & 56.4 \\
    FastQAExt\cite{weissenborn2017making} & 43.7 & 56.1 \\
    \bfseries{EvolutionRC} & \bfseries{47.0} & \bfseries{62.9} \\
    AMANDA\cite{kundu2018question} & 48.4 & 63.3 \\
    \hline
    \end{tabular}
    \end{center}
    \caption{\label{tab} The performance of EvolutionRC and competing models on NewsQA development set.}
\end{table*}

\section{5. Discussion}

There are two key points in our proposed framework: initializing population from existing models, and guiding the evolution process with a value-function based performance estimator.
To understand how they affect the process of evolution algorithm, We conduct ablation studies on these components of the proposed framework, including the proposed initialization method, and the performance estimator.

\subsection{5.1 Initialization Method}
To understand how the initialization method affect the final result of evolution algorithm, we run the framework on SQuAD dataset again, with completely random initialization. All other experiment setup is kept the same. Here is the table showing the difference. With the proposed initialization method, the EM score increases from 76.0 to 78.9.

\begin{table}[H]
    \begin{center}
    \begin{tabular}{|l|c|c|}
    \hline \bf Single Model & \bf Exact Match & \bf F1 \\ \hline
    EvolutionRC & 78.9 & 86.1 \\
    Naive Evolution\cite{Real2017Evolution} & 76.0 & 84.1 \\
    \hline
    \end{tabular}
    \end{center}
    \caption{\label{tab} The effect of initializing population with known models.}
\end{table}

\subsection{5.2 Performance Estimator}

To investigate how the CNN-based performance estimator affects the training procedure, we perform experiments on SQuAD dataset, with and without the performance estimator. When running this experiment, we initialize the population completely by random. The evolution process stops after 400 models are trained. We choose $epoch_{max} = 100$ for the performance estimator in this experiment.

\begin{figure}[H]
    \centering
    \includegraphics[scale=0.3]{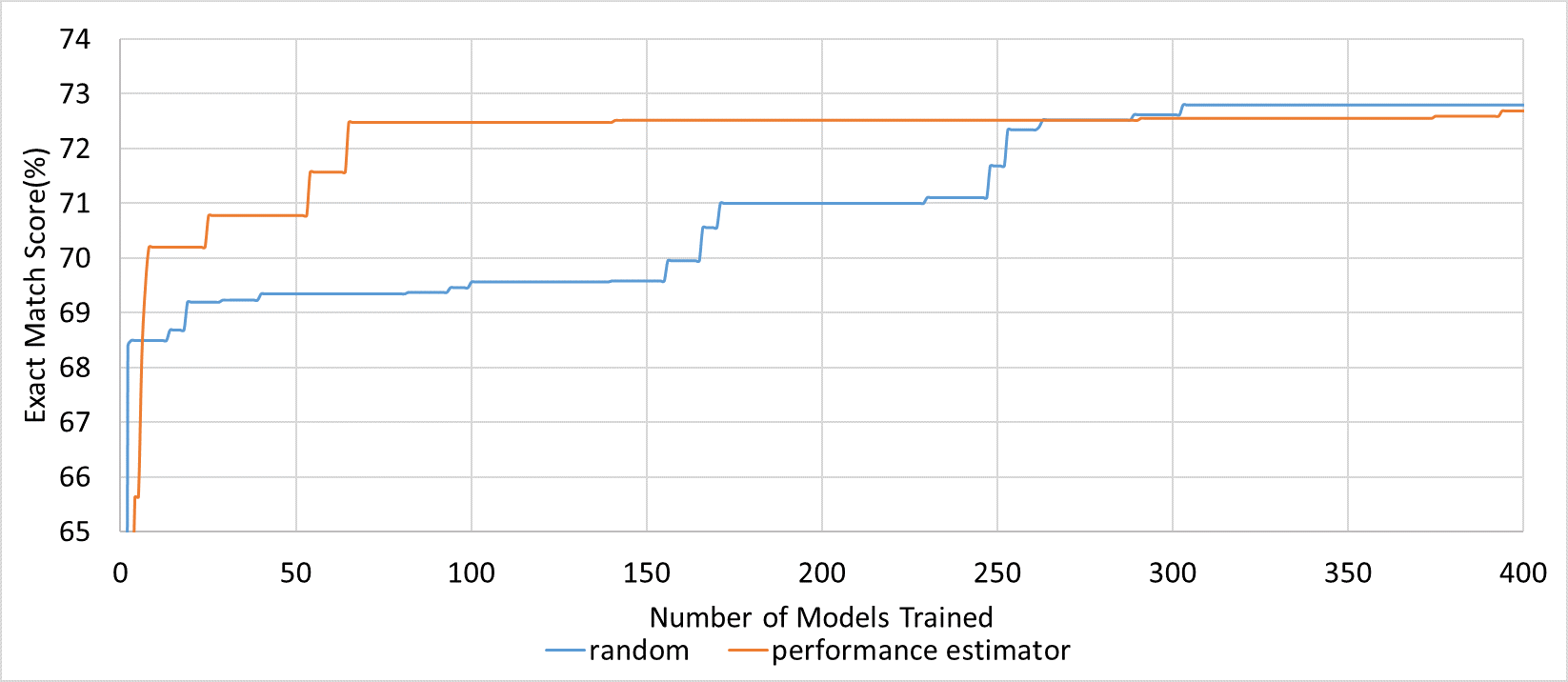}
    \caption{Convergence of evolution process}
\end{figure}
From the above figure, we can see the performance estimator speeds up the convergence of the evolution algorithm by avoiding ineffective mutations. However, the performance estimator does not improve the final result, so we didn't use it in the experiments mentioned in Section 4. The experiment is repeated only once, so it might be affected by random factors.

\section{6. Conclusion}
In this paper, we propose a new evolution algorithm based framework for different Question-Answering problems. We propose a new variant of evolution algorithm, which uses a CNN-based performance estimator, and existing human-designed models as starting points for the mutation process. Our experiment shows that (i) This framework achieves near state-of-art performance on multiple datasets with limited human intervention and acceptable computational resources. (ii) The proposed way of initializing population is better than random initialization, in terms of final result. (iii) The use of the proposed performance estimator speeds up the convergence of the search progress.

{\fontsize{9.0}{10.0}\selectfont
\bibliography{main}}
\end{document}